\documentclass{llncs}
\usepackage[T1]{fontenc}
\usepackage[utf8]{inputenc}
\usepackage{graphicx}
\usepackage{subcaption}
\usepackage{amsmath,amssymb}
\usepackage{booktabs}
\usepackage{multirow}
\usepackage{float}
\usepackage{xcolor}
\usepackage{hyperref}
\usepackage{marvosym}
\usepackage{bm}
\usepackage{amsmath}  
\usepackage{marvosym} 
\usepackage{hyperref} 
\usepackage[T1]{fontenc}
\usepackage{graphicx}
\begin{document}
\title{A Triple-Modal Contrastive Learning Framework with Sequence, Graph, and 3D Features for Drug–Target Interaction Prediction}
%
%
\author{
Le Xu \and Xi Zhang \and Dan Luo \and
Ting Wang\textsuperscript{(\Letter)} \and 
Xuan Lin\textsuperscript{(\Letter)}
}

\institute{
School of Computer Science, Xiangtan University, Xiangtan 411105, China\\
\email{jack\_lin@xtu.edu.cn}\\
}
\maketitle              
\begin{abstract}
Accurate prediction of drug-target interactions (DTI) is critical for drug discovery. Existing methods often rely on single-modal representations (e.g., sequences or graphs) or combine only two modalities, overlooking 3D structural features.To address this challenge, we propose TriMod-DTI, a triple-modal contrastive learning framework that incorporates 1D sequences, 2D graphs, and 3D structures of drugs and proteins, obtaining the universal and complementary feature represetations for DTI prediction. We design a Feature Extractor to capture drug and target features across the three modalities, thereby enriching their representations. We further propose a triple-modal contrastive learning strategy to align different modal representations of the same drug or protein in the latent space. By constructing cross-modal positive and negative sample pairs, this approach enhances the model’s discriminative ability. Experiments on three benchmark datasets demonstrate that TriMod-DTI outperforms state-of-the-art methods. The ablation studies validate the contributions of each modality. Moreover, case studies highlight its practical potential for DTI prediction and drug discovery. The source code of TriMod-DTI is available at \url{https://github.com/klez1/TriMod-DTI}.

\end{abstract}
\section{Introduction}
Accurate prediction of drug–target interactions (DTIs) is crucial for drug discovery, facilitating the identification of therapeutic targets and candidate compounds \cite{zhao2021biomedical}. Traditionally, DTIs are determined through experimental approaches such as high-throughput screening \cite{bajorath2002integration}. However, these methods are often expensive, time-consuming, and difficult to scale for large-scale drug discovery. 

In recent years, deep learning has made significant progress in DTI prediction \cite{mak2024artificial}. Existing deep learning models can be broadly categorized into unimodal and multimodal approaches. Unimodal methods model drugs and targets using information from a single modality. For example, TransformerCPI \cite{chen2020transformercpi} utilizes the Transformer architecture to represent drug and protein sequences. GraphDTA \cite{nguyen2021graphdta} leverages graph neural networks (GNNs) \cite{scarselli2008graph} to learn molecular graph representations.
AttentionSiteDTI \cite{yazdani2022attentionsitedti} focuses
on constructing drug and protein binding site
representations. Mutual-DTI \cite{wen2023mutual} incorporates a
multi-head self-attention mechanism to effectively
capture drug-protein interaction features, significantly improving prediction accuracy. MGMA-DTI \cite{li2025mgma} uses graph convolutional networks (GCNs) \cite{kipf2016semi} and multi-order gated convolution to encode drug and protein features, with a multi-attention fusion network to model their interactions.

Unimodal approaches have achieved promising results in DTI prediction, but they still have certain limitations. These models rely on singlemodality data, making it challenging to fully capture the complexity of drug-target interactions. Researchers have started exploring multi-modal
representations of drugs and proteins. IIFDTI \cite{cheng2022iifdti} integrates drug sequence and graph information with protein sequences, while CSCL-DTI \cite{lin2024cscl} enhances drug representation through sequence–graph fusion and contrastive learning. 
Despite their promising performance, most existing multimodal models simply fuse features from different modalities without explicitly modeling their relationships, which may lead to information redundancy or conflicts.

\begin{figure*}[!h]
\centering
\includegraphics[width=1\linewidth]{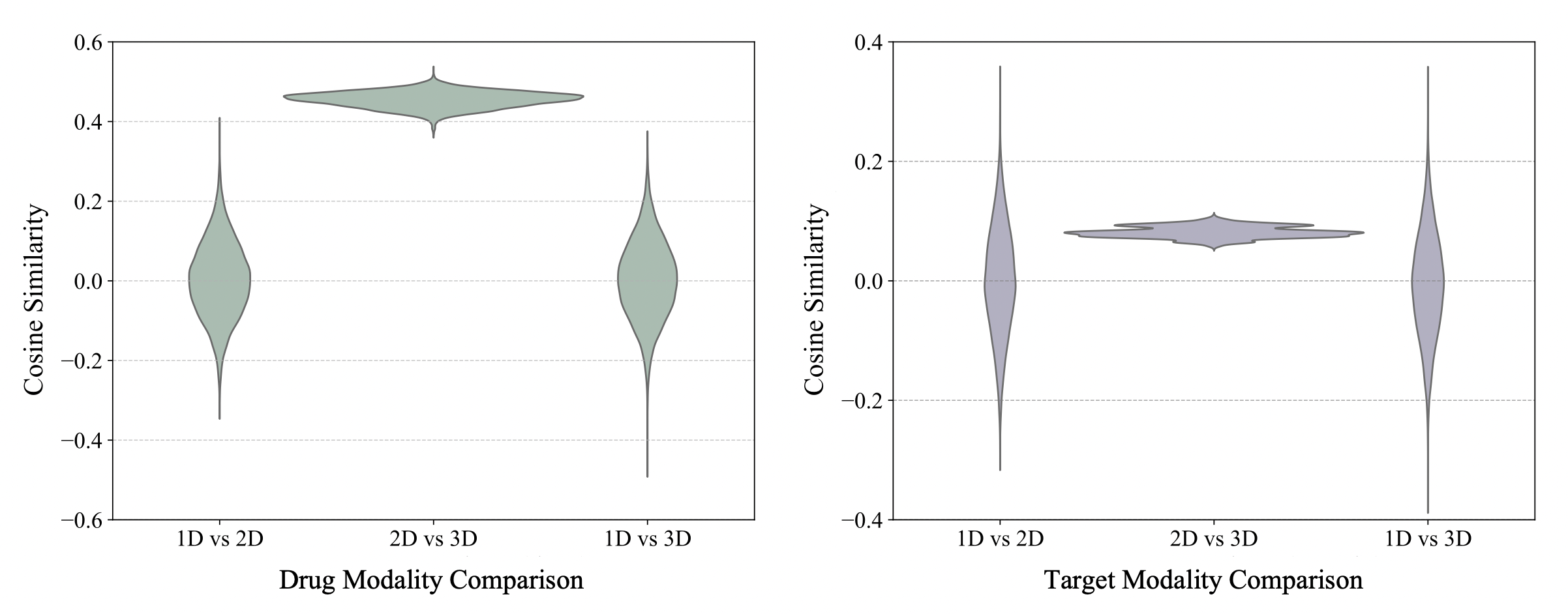}
\caption{The cosine similarity between the embeddings of different modalities.}
\label{fig:1}
\end{figure*}

To better understand the relationships among modalities, we conduct a cosine similarity analysis on embeddings derived from 1D sequences, 2D molecular graphs, and 3D structural representations of drugs and proteins on the GPCR dataset. As shown in Figure~\ref{fig:1}, similarities between modalities are generally low, mostly distributed within [$-0.25$, 0.25], indicating strong complementarity. These findings highlight the limitations of single-modality representations and suggest that effectively modeling cross-modal relationships is essential for improving DTI prediction.

To address these challenges, we propose TriMod-DTI, a framework that integrates sequence, graph, and 3D structural representations of drugs and proteins. Building upon CSCL-DTI \cite{lin2024cscl}, the model further incorporates protein binding-site graphs and 3D structures to enrich multimodal representations. In addition, we introduce a cross-modal contrastive learning mechanism inspired by the CLIP framework \cite{radford2021learning} to align multimodal embeddings. The main contributions of this work are summarized as follows.

\begin{itemize}
\item We propose TriMod-DTI, a multimodal DTI prediction framework that integrates sequence, graph, and 3D structural representations of drugs and proteins, and introduces a cross-modal contrastive learning strategy to align and fuse representations from different modalities.
\item Extensive experiments on three benchmark datasets demonstrate that TriMod-DTI consistently outperforms several state-of-the-art methods, while ablation studies further validate the effectiveness of each modality and the proposed contrastive learning mechanism.
\end{itemize}

\section{Materials and Methods}
\subsection{Datasets}
We conducted a comprehensive evaluation of the TriMod-DTI on three public datasets, namely GPCR, Human, and DrugBank. We used the RDKit \cite{landrum2013rdkit} to generate drug structure files and employed the OmegaFold \cite{wu2022high} to predict the 3D structural information of proteins. We utilized the dataset from previous studies \cite{lin2024cscl}. Due to the absence of 3D structural information for certain proteins and drug molecules, we performed data cleaning and removed the incomplete records, with the final dataset summarized in Table~\ref{tab:1}.

\begin{table}[t]
\centering
\caption{Details of datasets used in this work.}
\label{tab:1}
\begin{tabular}{cccccc}
\toprule
Datasets  & Drugs & Targets & Interactions & Positive & Negative\\
\midrule
Human & 2,175 & 1,540 & 4,965 & 2,330  & 2,635\\
GPCR & 4,976 & 330 & 14,376 & 7,479 & 6,897\\
DrugBank & 1,441 & 1,262 & 13,121 & 5,634 & 7,487\\
\bottomrule
\end{tabular}
\end{table}

\subsection{Methodological Overview}
\begin{figure*}[!h]
\centering
\includegraphics[width=1\linewidth]{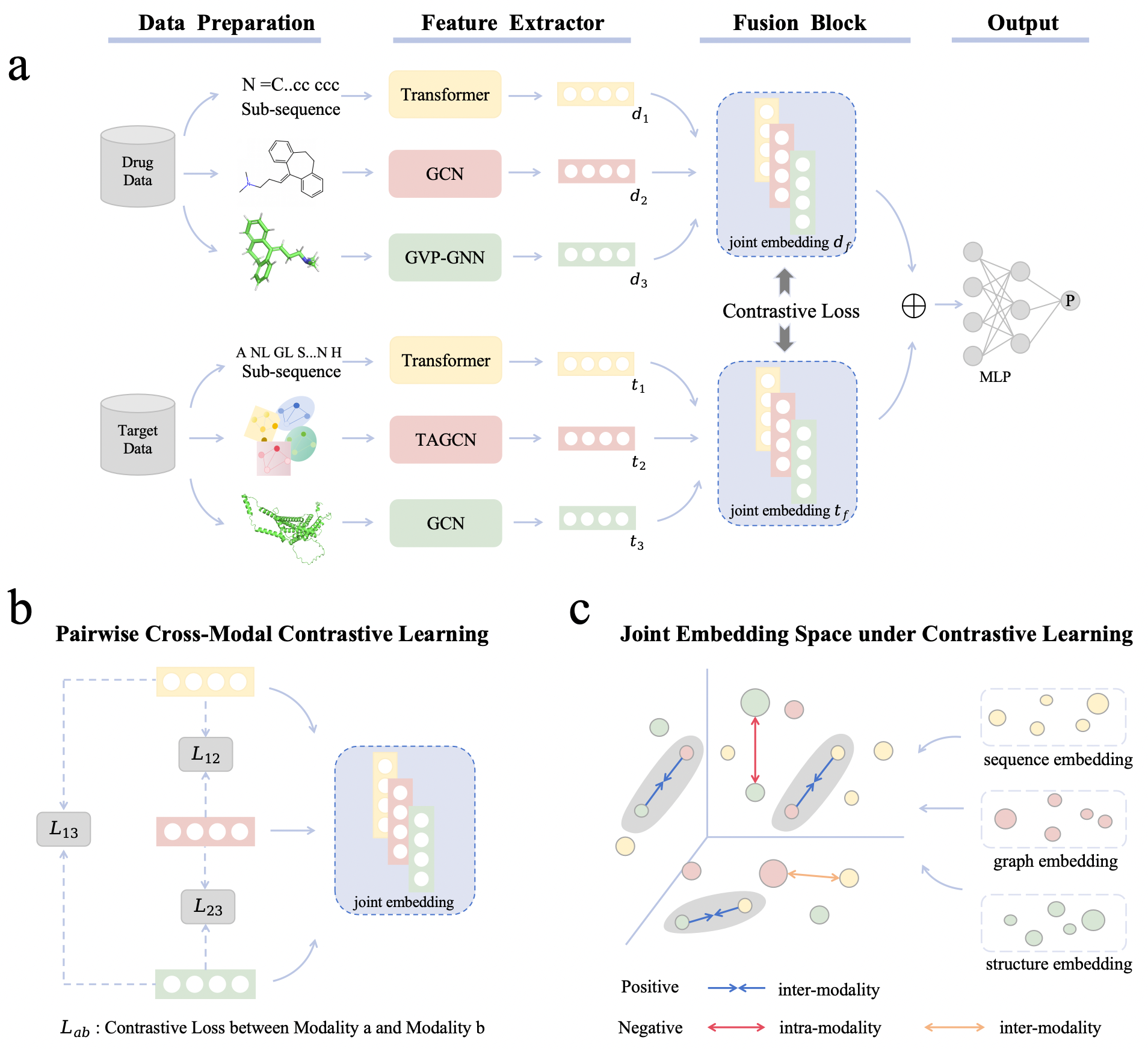}
\caption{Illustration of TriMod-DTI. (a) The framework of proposed TriMod-DTI.(b) Details of cross-modal contrastive learning among three modalities in the Fusion Block of architecture (a). (c) Joint feature vector space of three Modalities under cross-modal contrastive learning in (b).}
\label{fig:2}
\end{figure*}

An overview of the TriMod-DTI framework is illustrated in Figure~\ref{fig:2}. The DTI prediction task is formulated as a binary classification problem by leveraging multi-modal representations of drugs and targets. The framework comprises three sequential stages, where sequence, graph, and structural information are extracted for both drugs and proteins, modality-specific encoders are employed to learn feature embeddings from each individual modality, and the learned drug and target representations are fused to construct a joint feature representation that is fed into a classifier for DTI prediction.

\subsection{Feature Extraction for Drug}
We adopt a multimodal framework to represent drugs by integrating sequence, graph, and structural information.

\noindent \textbf{Sequence representation.}
Drug SMILES sequences are segmented into sub-sequences using the FCS algorithm \cite{huang2021moltrans}. Each substructure is mapped to a positional embedding ${E}{pos_i}^d$ and a content embedding ${E}{con_i}^d$, and the combined embedding is defined as
\begin{equation}
{E}{i}^{d} = {E}{pos_{i}}^d + {E}{con{i}}^d.
\end{equation}
The embedding matrix ${E}^{d}\in \mathbb{R}^{l_d \times \vartheta}$ is then processed by a Transformer encoder \cite{vaswani2017attention} to capture contextual relationships between substructures, producing the sequence representation
\begin{equation}
\tilde{E}^{d}={Transformer}_{d}({E}^{d}).
\end{equation}

\noindent \textbf{Graph representation.}
We employ RDKit \cite{landrum2013rdkit} to convert drug SMILES into molecular graphs $G_{d_2}(\mathcal{V},\mathcal{E})$, where $\mathcal{V}$ and $\mathcal{E}$ denote the sets of atoms and covalent bonds, respectively. Each atom is represented by a 75-dimensional feature vector constructed from one-hot encodings of atomic properties. The node feature matrix is denoted as $M_d \in \mathbb{R}^{N_a \times 75}$, where $N_a$ represents the number of atoms, and $A$ denotes the adjacency matrix of the molecular graph.

The molecular graph is encoded by a GCN \cite{kipf2016semi} to obtain the drug representation $d_2$ according to
\begin{equation}
Z_d^{i+1} = \sigma(GCN(\tilde{A}, W^{i}, b^{i}, Z_d^{i})),
\end{equation}
where $\tilde{A}$ denotes the self-loop adjacency matrix, $\sigma$ represents the nonlinear activation function, and the initial representation $Z_d^{0}$ is obtained by projecting $M_d$ into a $D_a$-dimensional embedding space.

\noindent \textbf{Structural representation.}
To incorporate spatial information, we construct a 3D molecular graph $G_{d_3}=(\mathcal{V},\mathcal{E})$ from RDKit-generated SDF files. Edges are established between atoms whose Euclidean distance is smaller than 4.5 Å \cite{mu2024medication}. Node and edge features encode atomic coordinates and bond information. Geometric Vector Perceptrons (GVP) \cite{jing2020learning} are employed to jointly process scalar and vector features according to
\begin{equation}
s' = \sigma(W_v \, Concat(\|W_h V\|_2 , s) + b),
\end{equation}
\begin{equation}
V' = \sigma(\|W_u W_h V\|_2) \odot \|W_u W_h V\|_2 .
\end{equation}

A GVP-GNN \cite{jing2020learning} further updates node representations through message passing
\begin{equation}
m_{j \rightarrow i}^{(l)} = g(Concat(v_j^{(l-1)}, \epsilon_{j \rightarrow i})),
\end{equation}
\begin{equation}
v_i^{(l)} = LayerNorm\left(v_i^{(l-1)} + \frac{1}{|N(i)|}\sum_{j \in N(i)} m_{j \rightarrow i}^{(l)}\right).
\end{equation}

Finally, global add pooling aggregates node embeddings to obtain the 3D molecular representation, whose dimensionality is aligned with the sequence and 2D graph representations for subsequent cross-modal contrastive learning.

\subsection{Feature Extraction for Target}
To obtain informative target representations, we extract sequence, binding-site graph, and 3D structural features of proteins.

\noindent \textbf{Sequence representation.}
Protein sequences are first segmented using the FCS algorithm. Each amino acid is mapped to a content embedding ${E}{con_i}^{p}$ and a positional embedding ${E}{pos_i}^{p}$. The initial embedding is defined as
\begin{equation}
{E}{i}^{p} = {E}{pos_{i}}^{p}+{E}{con_{i}}^{p}.
\end{equation}
The embedding matrix $E^{p}\in \mathbb{R}^{l_p \times \vartheta}$ is fed into a Transformer encoder to obtain the protein sequence representation $t_1$.

\noindent \textbf{Graph representation.}
Protein 3D structures are predicted using OmegaFold \cite{wu2022high}. Binding sites are identified following the method of Saberi Fathi et al. \cite{saberi2014simple}. Each pocket is represented as a graph, where nodes denote atoms and edges represent atomic connections. Following AttentionSiteDTI \cite{yazdani2022attentionsitedti}, each atom is encoded as a 31-dimensional feature vector containing atom type, degree, hydrogen count, and implicit valency.

Since a protein may contain multiple binding pockets, multiple pocket graphs are constructed. These graphs are encoded using a Topology Adaptive Graph Convolutional Network (TAGCN) \cite{du2017topology}. The convolution operation is defined as
\begin{equation}
H=\sum_{k=1}^{K}(D^{-1/2}AD^{-1/2})^k X W_k + b ,
\end{equation}
where $A$ and $D$ denote the adjacency and degree matrices, and $X$ represents the initial node features. After graph convolution, global attention pooling aggregates node features within each pocket to obtain pocket-level embeddings. The final protein representation $t_2$ is obtained through a fully connected layer.

\noindent \textbf{Structural representation.}
For 3D structural modeling, the coordinates of $C_{\alpha}$ atoms and residue types are extracted from the PDB file. Residues are encoded as node features, and edges are constructed using the NeighborSearch function in Biopython \cite{cock2009biopython} with an 8Å threshold \cite{you2022cross}. The resulting graph is processed using a GCN encoder to obtain the structural representation $t_3$.

\subsection{Fusion of Drug and Target Representations}
We introduce a Fusion Block to integrate the three modalities. Because representations from different modalities may exhibit distribution discrepancies, direct fusion can cause information loss. To mitigate this issue, a tri-modal alignment strategy based on contrastive learning is employed. A cross-modal contrastive loss aligns representations of the same drug (or protein) while separating those of different samples in the embedding space. As illustrated in Figure~\ref{fig:2}c, cross-modal pairs of the same drug are treated as positive samples, whereas intra- and cross-modal pairs from different drugs are regarded as negative samples. The same strategy is applied to proteins.
The contrastive loss for drugs is defined as
\begin{equation}
\mathcal{L}_{\text{CL}}^d=\frac{1}{3}(\mathcal{L}_{12}^d+\mathcal{L}_{23}^d+\mathcal{L}_{13}^d),
\end{equation}
where $\mathcal{L}_{ab}^d = \mathcal{L}_{\text{CL}}^a + \mathcal{L}_{\text{CL}}^b$, and
\begin{equation}
\mathcal{L}_{\text{CL}}^a = -\frac{1}{2}\sum_{i=1}^{N_d}\log\frac{\exp(s(d_i^a, d_i^b)/\tau)}{\sum_{j=1}^{N}\left[\exp(s(d_i^a, d_j^b)/\tau) + \exp(s(d_i^a, d_j^a)/\tau)\right]},
\end{equation}
with $\mathcal{L}_{\text{CL}}^b$ following the same form as $\mathcal{L}_{\text{CL}}^a$ (swapping $a$ and $b$). Here, $\tau$ is the temperature parameter, $N$ is the batch size, and $s(\cdot,\cdot)$ denotes cosine similarity. The contrastive loss for targets is defined identically. After the alignment process, the representations derived from various modalities are concatenated together to construct the joint feature vector, which is defined as follows.
\begin{equation}
F = (d_1 \oplus d_2 \oplus d_3) \oplus (t_1 \oplus t_2 \oplus t_3).
\end{equation}

\subsection{Drug-Target Interaction Prediction}
In this study, the fused features are passed through three fully connected layers to map them to the final classification output. We employ an MLP with ReLU activation in the hidden layers and a Sigmoid activation in the output layer for prediction. We optimize the model using cross-entropy loss that is mathematically expressed as follows.
\begin{equation}
\mathcal{L}(y, \hat{y}) = -\frac{1}{N} \sum_{i=1}^{N} \left[ y_i \log(\hat{y}_i) + (1 - y_i) \log(1 - \hat{y}_i) \right],
\end{equation}
where $N$ is the number of training samples. $y_i$ and $\hat{y}_{i}$ denote known and predicted labels respectively. The total loss comprises classifier and contrastive loss.
\begin{equation}
\mathcal{L} = \alpha\mathcal{L}(y, \hat{y})+\beta\mathcal{L}_{CL}^d+\gamma\mathcal{L}_{CL}^p,
\end{equation}
where $\alpha$, $\beta$, and $\gamma$ are hyperparameters that control the influence of each loss on the model's training process.

\section{Results and Analysis}
\subsection{Experimental Setting and Evaluation Metrics}
Following existing works \cite{cheng2022iifdti,lin2024cscl}, we adopted different dataset splitting strategies for DTI evaluation.For the Human and DrugBank datasets, we used five-fold cross-validation, randomly dividing the dataset into training, validation, and test sets with a ratio of 8:1:1.  For the GPCR dataset, 20\% of the training set was used as validation.Each dataset split was evaluated over 10 independent runs to ensure reliability of the results.  Implementation was based on PyTorch 2.1.0 on an Ubuntu 22.04 machine with an RTX 4090 (24GB) GPU. The model with the best validation AUC was used for testing.

We select widely recognized evaluation metrics, including the area under the receiver operating characteristic curve (AUC), area under the precision-recall curve (AUPR) and Precision as our primary evaluation criteria. 

\subsection{Performance Comparison with Baselines}
We evaluated the effectiveness of the proposed model by conducting a detailed comparison with six baseline methods. Among them, TransformerCPI\cite{chen2020transformercpi}, Mutual-DTI\cite{wen2023mutual} and MGMA-DTI focus solely on the sequence information of drugs and proteins, while GraphDTA\cite{nguyen2021graphdta}, IIFDTI\cite{cheng2022iifdti} and CSCL-DTI\cite{lin2024cscl} consider both the sequence and graph structural information of drug-target interactions. The training protocols for all baseline models were kept consistent with those of TriMod-DTI to ensure a fair comparison.

\begin{table}[!htb]
\captionsetup{justification=raggedright, singlelinecheck=false}
\centering
\caption{Comparison results of TriMod-DTI and baselines on three datasets.}
\label{tab:2} 
\begin{tabular}{ccccc}
\toprule
Dataset & Method & AUC & AUPR & Precision \\
\midrule
\multirow{7}{*}{Human} & TransformerCPI & $0.928 \pm 0.070$ & $0.929 \pm 0.074$ & $0.832 \pm 0.089$ \\
& GraphDTA & $ 0.947 \pm 0.046 $ & $ 0.943 \pm 0.042$ & $0.884 \pm 0.087 $\\
& IIFDTI & $ 0.980 \pm 0.001 $ & {$ 0.982 \pm 0.003$}& $ 0.914 \pm 0.036 $\\
& Mutual-DTI & $ 0.973 \pm 0.005$ & $ 0.976\pm0.005$ & $0.907\pm0.015$ \\
& CSCL-DTI & {$ 0.981 \pm 0.003 $} & $ 0.981\pm0.005$ & {$0.943 \pm 0.031$}\\
& MGMA-DTI & $0.981\pm0.003$ & $0.981\pm0.004$ & $0.942\pm0.021$ \\
& \textbf{TriMod-DTI} & $\bm{0.988\pm0.003}$ & $\bm{0.986\pm0.004}$ & $\bm{0.945\pm0.022}$ \\
\midrule
\multirow{7}{*}{GPCR} & TransformerCPI & $0.804\pm0.054$ & $0.801\pm0.058$ & $0.697\pm0.053$ \\
& GraphDTA & $0.809\pm0.033$ & $0.802\pm0.034$ & $0.702\pm0.035$ \\
& IIFDTI & $0.825\pm0.008$ & $0.822\pm0.007$ & $0.714\pm0.025$  \\
& Mutual-DTI & $0.766\pm0.002$ & $0.786\pm0.005$ & $0.672\pm0.006$  \\
& CSCL-DTI & {$0.852\pm0.005$} & {$0.841\pm0.012$} & {$0.739\pm0.008$}  \\
& MGMA-DTI & $0.868\pm0.025$ & $0.852\pm0.019$ & $0.744\pm0.004$ \\
& \textbf{TriMod-DTI} & $\bm{0.870 \pm 0.015}$  & $\bm{0.868\pm0.002}$ & $\bm{0.751\pm0.010}$  \\
\midrule
\multirow{7}{*}{DrugBank} & TransformerCPI & $0.772\pm0.059$ & $0.520\pm0.102$ & $0.402\pm0.079$ \\
& GraphDTA & $0.808\pm0.024$ & $ 0.551\pm0.042 $ & $0.445\pm0.094$  \\
& IIFDTI & $0.805\pm0.013$ & $0.521\pm0.021$ & $0.426\pm0.010$ \\
& Mutual-DTI & $0.799\pm0.009$ & {$0.550\pm0.012$} & $0.431\pm0.012$  \\
& CSCL-DTI & $0.815\pm0.008$ & $0.549\pm0.027$ & $0.446\pm0.016$  \\
& MGMA-DTI & $0.812\pm0.006$ & $\bm{0.551\pm0.012}$ & $0.425\pm0.024$\\
& \textbf{TriMod-DTI} & $\bm{0.818\pm0.007}$ & $0.532\pm0.018$ & $\bm{0.463\pm0.011}$  \\
\bottomrule
\end{tabular}
\end{table}

Table~\ref{tab:2} compares the performance of different models on three datasets. On the Human dataset, TriMod-DTI slightly outperforms the baselines, improving AUC and AUPR by 0.7\% and 0.5\% over the second-best method. On the GPCR dataset, it shows more notable gains, surpassing MGMA-DTI \cite{li2025mgma} by 0.2\% in AUC, 1.6\% in AUPR, and 0.7\% in Precision. On the DrugBank dataset, although TriMod-DTI achieves the best AUC and Precision, its AUPR is relatively lower, possibly due to the stronger class imbalance, which leads the model to predict positives only at high confidence.

\subsection{Ablation Study}

We conducted ablation studies to evaluate the contrastive learning strategy and the contribution of each modality. Due to space limitations, detailed results are provided in our GitHub repository. For contrastive learning ablation, removing the full contrastive loss ($\mathcal{L}_{CL}$) or any of its cross-modal components ($\mathcal{L}_{12}$, $\mathcal{L}_{23}$, $\mathcal{L}_{13}$) consistently degraded performance. The full contrastive learning achieved gains of 1.1\% in AUC and 2.0\% in AUPR over the non-contrastive baseline.

\begin{figure*}[!h]
\centering
\includegraphics[width=1\linewidth]{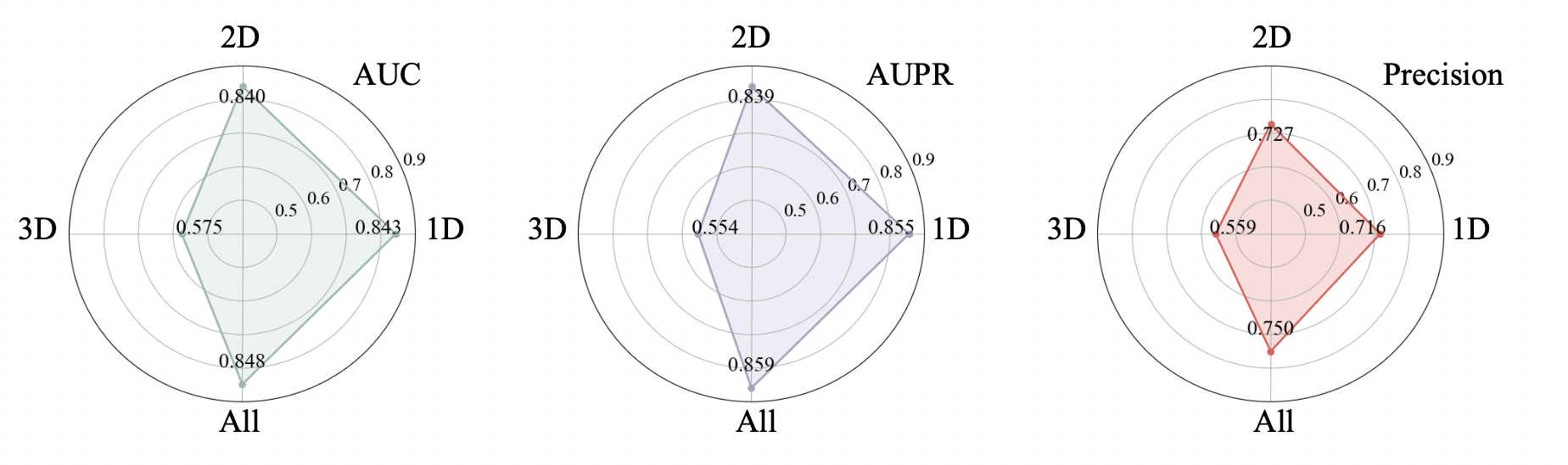}
\caption{Performance of each modality encoded independently on the GPCR.}
\label{fig:3}
\end{figure*}

As shown in Figure~\ref{fig:3}, the sequence modality contributes most to performance, followed by the graph modality, while the 3D modality alone performs relatively poorly. This may be attributed to our 3D encoder design, which primarily incorporates atomic coordinates and bond information but excludes essential chemical properties such as atom types, degrees, and formal charges. Nevertheless, integrating all three modalities enables complementary learning: sequence and graph features capture global molecular information, while the 3D structure provides local spatial context. This integration leads to the best overall performance of TriMod-DTI.

\begin{figure*}[!h]
\centering
\includegraphics[width=1\linewidth]{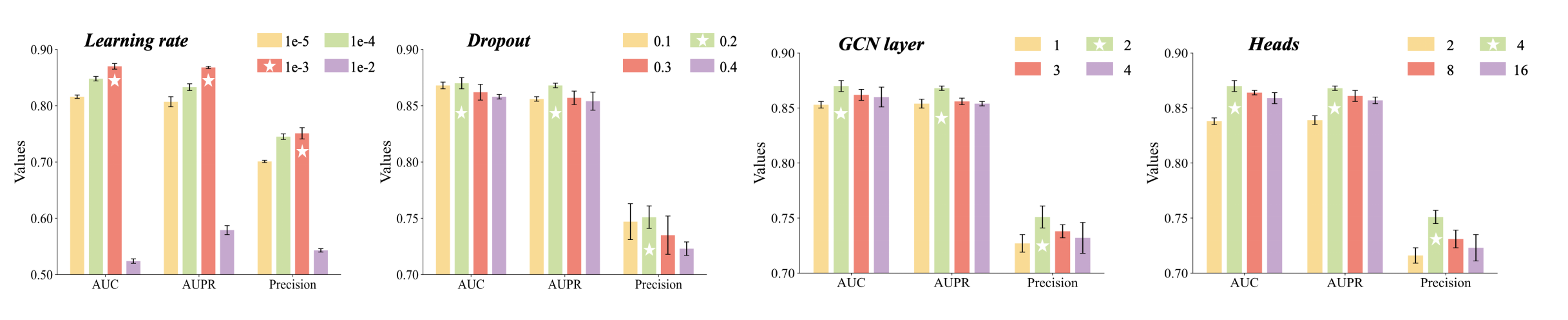}
\caption{Parameter sensitivity analysis for TriMod-DTI on GPCR dataset.}
\label{fig:4}
\end{figure*}

\subsection{Parameter Sensitivity Analysis}
We systematically investigated four key parameters in TriMod-DTI, including the dropout rate, learning rate, number of GCN layers, and number of cross-attention heads. As shown in Figure~\ref{fig:4}, on the GPCR dataset, the model achieved optimal performance with dropout rate of 0.2, learning rate of 1e-3, 2 GCN layers, and 4 attention heads. Detailed sensitivity analysis results on the Human dataset are available in our GitHub repository.

\subsection{Case Study}
\begin{figure*}[!h]
\centering
\includegraphics[width=1\linewidth]{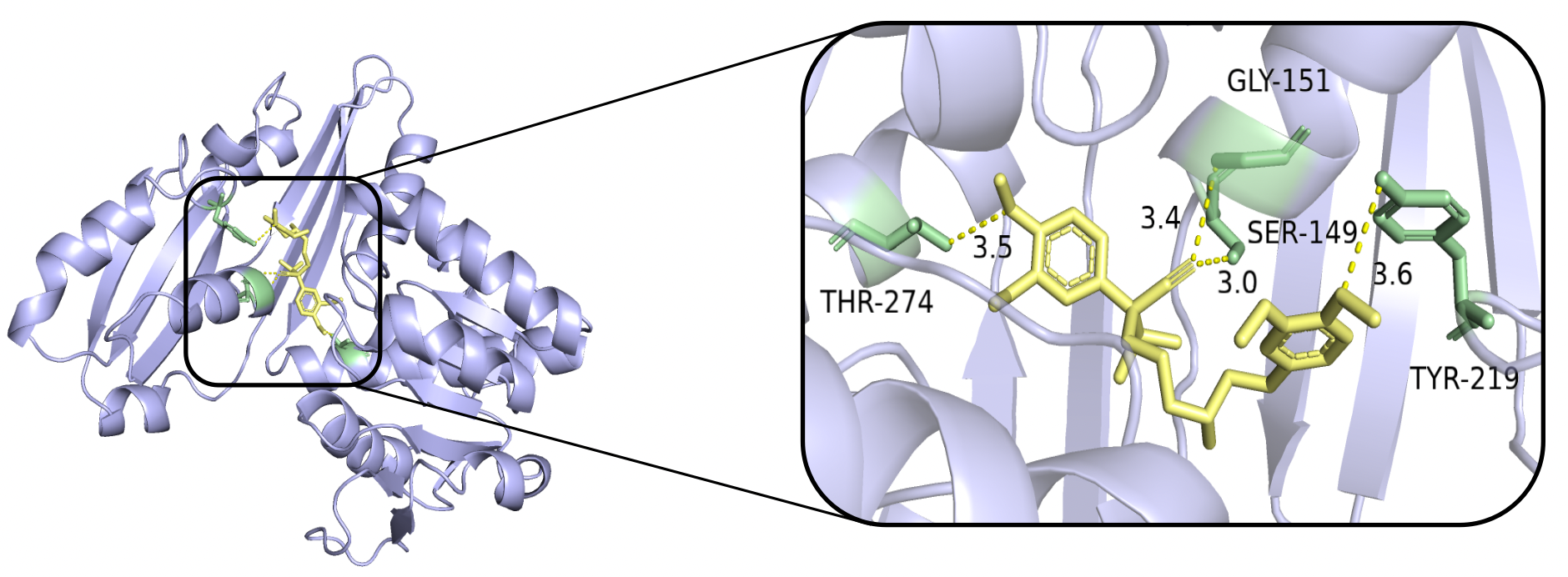}
\caption{Visualization of drug-protein interactions.}
\label{fig:5}
\end{figure*}

To evaluate TriMod-DTI, we conducted a case study on Verapamil (DB00661) and D(2) dopamine receptor (P14416). From the top-10 predicted candidate drugs for the receptor shown in Table~\ref{tab:4}, 5 were validated by literature on PubMed\footnote{\url{https://pubmed.ncbi.nlm.nih.gov}}. Furthermore, we performed molecular docking between Verapamil and its top-ranked predicted target (Glucose-6-phosphate isomerase 2, P13376) using the DeepMice\footnote{\url{http://www.deepmice.com}} server. Visualization with PyMOL\footnote{\url{https://www.pymol.org}} as presented in Figure~\ref{fig:5} showed that Verapamil forms hydrogen bonds with key residues (THR-274, GLY-151, SER-149) within the binding pocket, stabilizing the complex and providing structural evidence for the predicted interaction. Full target prediction results are available in our GitHub repository.

\begin{table}
\centering
\caption{The predicted candidate targets for drug \textit{Verapamil}.}\label{tab:4}
\begin{tabular}{cccc}
\hline
Rank & Target Name & Target Uniprot ID & Evidence\\
\hline
1 & Meso-diaminopimelate D-dehydrogenase & P04964 & Unconfirmed \\
2 & Adenylate kinase & P69441 & Unconfirmed \\
3 & \textbf{Glucose-6-phosphate isomerase 2} & \textbf{P13376} & \textbf{PMID: 36978873} \\
4 & Purine nucleoside phosphorylase & P00491 & Unconfirmed\\
5 & \textbf{Heparin cofactor 2} & \textbf{P05546} & \textbf{PMID: 31713254} \\
6 & \textbf{Proteasome subunit alpha type-5} & \textbf{P28066} & \textbf{PMID: 27813130} \\
7 & Ras-related protein Ral-A & P11233 & Unconfirmed \\
8 & \textbf{Cytochrome b6-f complex subunit 8} & \textbf{P83798} & \textbf{PMID: 37831396} \\
9 & \textbf{Ornithine decarboxylase antizyme 1} & \textbf{P54368} & \textbf{PMID: 2491756} \\
10 & Dihydropteridine reductase & P09417 & Unconfirmed \\
\hline
\end{tabular}
\end{table}

\section{Discussion}
This study presents TriMod-DTI, a multimodal framework for drug–target interaction prediction that integrates sequence, graph, and structural representations of drugs and targets. By employing modality-specific encoders and a cross-modal contrastive learning strategy, the model effectively aligns heterogeneous representations and captures complementary information across modalities. Experimental results on three benchmark datasets demonstrate that TriMod-DTI achieves competitive performance compared with existing methods and shows potential in supporting drug discovery tasks. In the future, more advanced geometric learning techniques and modality-aware contrastive strategies can be explored to further enhance the performance of multimodal representation learning.

\begin{credits}
\subsubsection{\ackname} The work is supported in part by the National Natural Science Foundation of China (Grant No.62573372), in part by the Hunan Province College Students’ Innovation
Training Program (Grant No. S202410530023),and the Hunan
Provincial Key Research and Development Program Project (Grant No.2025JK2003).

\subsubsection{\discintname}
The authors have no competing interests to declare that are
relevant to the content of this article.
\end{credits}
%

\end{document}